\documentclass[letterpaper]{article}

\usepackage{natbib,alifeconf}  
\usepackage{url,hyperref}
\usepackage{booktabs}
\usepackage{amsmath,amssymb}
\usepackage{graphicx}
\usepackage{float}

\title{Telogenesis: Goal Is All U Need}

\author{Zhuoran Deng$^{1}$, Yizhi Zhang$^{1}$, Ziyi Zhang$^{1}$, Wan Shen$^{1}$ \\ \\
\textit{$^{1}$Independent Research}}

\begin{document}
\maketitle

\begin{abstract}
Goal-conditioned systems assume goals are provided externally. We ask whether attentional priorities---a minimal form of goal---can emerge endogenously from an agent's internal cognitive state. We propose a \textit{priority function} $\pi_i(t)$ that generates targets from three types of epistemic gap: ignorance (posterior variance), surprise (prediction error indicating model mismatch), and staleness (temporal decay of confidence in unobserved variables). We validate this mechanism in two systems: a minimal attention-allocation environment (2{,}000 Monte Carlo runs) and a modular, partially observable world of higher complexity (500 runs). Component ablation shows each term is necessary. A key finding emerges from contrasting two evaluation metrics: under \textit{global prediction error}, coverage-based rotation outperforms priority-guided allocation; under \textit{change detection latency}, the result reverses, with priority-guided allocation detecting environmental changes significantly faster, and this advantage growing monotonically with dimensionality ($d = -0.95$ at $N\!=\!48$, $p < 10^{-6}$). Detection latency follows a power law in attention budget, with priority-guided observation exhibiting a steeper exponent than rotation (0.55 vs.\ 0.40). When the staleness decay rate $\lambda$ is made learnable per variable, the system spontaneously recovers the volatility structure of the environment without supervision: learned $\lambda_i$ values differentiate high-volatility from low-volatility variables ($\bar{\lambda}_{\text{high}} = 0.289$ vs.\ $\bar{\lambda}_{\text{low}} = 0.202$, $t = 22.5$, $p < 10^{-6}$). We demonstrate that in attention-limited agents, epistemic gaps alone---without external reward---are sufficient to generate adaptive priority structures that outperform fixed strategies and spontaneously recover latent environmental structure.
\end{abstract}

\section{Introduction}

A central open problem in artificial intelligence is how autonomous agents generate their own goals. Goal-conditioned reinforcement learning \citep{schaul2015universal, andrychowicz2017hindsight} has demonstrated that agents can pursue diverse objectives, but these are invariably specified externally. The question of where goals \textit{come from} remains unaddressed.

Biological organisms do not receive goal specifications from an external reward function. An animal exploring a novel environment generates exploratory targets from its own internal state---directing attention toward aspects of the environment that are uncertain, surprising, or insufficiently modeled \citep{friston2015active, gottlieb2013information}.

We propose that attentional priority generation---a minimal but essential component of goal formation---can be formalized as an \textit{endogenous} process driven by epistemic gaps in the agent's world model. We do not claim full-fledged goal semantics; rather, we study attentional target formation as a minimal operational precursor of goal generation. We introduce a priority function that unifies three types of cognitive deficit into a single scalar score, from which targets emerge through softmax competition. We call this framework \textbf{Telogenesis} (from Greek \textit{telos}, purpose, and \textit{genesis}, origin): the origin of purpose from within.

A central methodological finding of this work is that the \textit{choice of evaluation metric} determines whether endogenous priority appears advantageous or not. Under global prediction error---a metric that assumes omniscient access to all state variables---coverage strategies are optimal. Under change detection latency---a metric available to the agent itself---priority-guided allocation dominates, with advantage scaling monotonically with environmental complexity. We argue that the latter is the appropriate measure of adaptation in partially observable worlds.

Our contributions are: (1)~a formal priority function decomposed into ignorance, surprise, and staleness; (2)~ablation in a minimal system demonstrating each component's necessity; (3)~validation in a modular partially observable environment; (4)~identification of a metric-dependent reversal in which strategy appears superior; (5)~a power law relating detection speed to attention budget; and (6)~demonstration that per-variable learned decay rates spontaneously recover latent environmental volatility structure without supervision.

\section{Related Work}

The priority function draws on several traditions while differing from each in specific ways.

\textbf{Intrinsic motivation and curiosity.} Schmidhuber's formal theory of fun \citep{schmidhuber2010formal} and curiosity-driven exploration \citep{pathak2017curiosity, burda2019exploration} use prediction error or learning progress to drive exploration. Our surprise term serves a similar role, but these methods operate in reward-augmented settings where curiosity bonuses supplement an external objective. Our framework uses no external reward at any level. Additionally, curiosity methods typically lack a staleness mechanism: they cannot generate priority for variables that have simply not been observed recently.

\textbf{Active inference.} Friston's active inference framework \citep{friston2015active} formalizes action selection as expected free energy minimization, encompassing both epistemic and pragmatic value. Our priority function can be viewed as a restricted form of epistemic value computation. The key difference is scope: active inference is a complete theory of perception and action, whereas we isolate the specific sub-problem of observation allocation and provide controlled experiments comparing endogenous priority against fixed baselines under partial observability.

\textbf{Bayesian experimental design.} Lindley's information gain \citep{lindley1956measure} and its descendants formalize optimal experiment selection. Our ignorance term (posterior variance) is closely related. However, classical experimental design assumes the ability to compute expected information gain over all candidate observations---a combinatorial computation that scales poorly. The priority function provides a computationally cheap approximation by combining three heuristic signals into a scalar score.

\textbf{Attention in partially observable environments.} Optimal attention allocation under partial observability has been studied in sensor scheduling \citep{hero2007sensor}, POMDP-based information gathering \citep{kaelbling1998planning}, and belief-space planning where actions are selected to reduce state uncertainty \citep{platt2010belief}. Our contribution is not a new algorithm for this problem but a specific claim: that the evaluation metric matters more than the allocation strategy, and that detection latency---not global error---is the appropriate measure of adaptation for the agent itself.

Our primary novelty is therefore not any individual component, but their unification into a single priority mechanism, the identification of metric-dependent reversal in strategy evaluation, and the demonstration that learned per-variable decay rates recover latent environmental structure without supervision.

\section{The Priority Function}

Consider an agent maintaining a Bayesian world model over $N$ environment variables, able to observe only $b \ll N$ per time step. We propose that observation selection emerges from an \textit{epistemic gap score}:

\begin{equation}
\pi_i(t) = w_1\,\tilde{\sigma}_i^2(t) + w_2\,\tilde{S}_i(t) + w_3\,(1 - e^{-\lambda \Delta t_i})
\label{eq:priority}
\end{equation}

\textbf{Ignorance} $\tilde{\sigma}_i^2(t)$: normalized posterior variance. High when data is insufficient. Decreases monotonically with observations but cannot detect that the world has changed since last observation.

\textbf{Surprise} $\tilde{S}_i(t)$: normalized prediction error, $S_i = |x_i - \hat{x}_i| / (\hat{\sigma}_i + \epsilon)$. Spikes when observations violate expectations, signaling model mismatch. Only available for observed variables.

\textbf{Staleness} $(1 - e^{-\lambda \Delta t_i})$: saturating function of time since last observation. Captures that confidence should decay for unobserved variables. The key innovation: this generates priority for variables \textit{without requiring any observation}, based purely on temporal reasoning.

Targets emerge via softmax: $P(\text{target}\!=\!i) = \exp(\pi_i/\tau) / \sum_j \exp(\pi_j/\tau)$. Temperature $\tau$ controls exploitation-exploration. An activation threshold $\theta$ provides dormancy in stable environments.

The function $\pi_i(t)$ is not an uncertainty measure but an \textit{information value function}. Surprise and staleness are not epistemic uncertainty per se; they are signals that make a variable \textit{worth attending to}. We use ``epistemic gap'' to encompass all three: gaps in knowledge, gaps between model and reality, and gaps in temporal coverage.

\section{Experiment 1: Minimal System}

\subsection{Design}

$N$ scalar variables in $[0,1]$, asymmetric observation noise (linearly spaced $\sigma \in [0.25, 0.05]$), $K < N$ variables change on regime switches every $T_r$ ticks, remainder stable. Agent observes one variable per tick, maintains Gaussian posteriors. Unobserved variables undergo variance inflation at rate $\gamma$/tick. Five groups: \textsc{Random} (uniform selection), \textsc{Priority} (full $\pi$), \textsc{Var-only} (ignorance term alone), \textsc{Rotation} (deterministic cycle), \textsc{Error} (greedy on last prediction error). Default: $N\!=\!6$, $K\!=\!3$, $T_r\!=\!15$, 2{,}000 runs.

\subsection{Component Ablation}

\textbf{v1 (Symmetric, pure variance):} No group differences. Without asymmetry, selective attention provides no advantage.

\textbf{v2 (Asymmetric, pure variance):} \textsc{Var-only} $\approx$ \textsc{Random}. Posterior variance cannot detect model mismatch.

\textbf{v3 (+ Surprise):} \textsc{Priority} shifts attention toward switching variables (53.2\% vs.\ 50.0\%), but only for observed variables.

\textbf{v4 (+ Staleness = Full $\pi$):} \textsc{Priority} significantly outperforms \textsc{Random} ($p < 10^{-6}$, $d\!=\!0.15$) and \textsc{Var-only} ($p < 0.001$). All three components jointly required.

\subsection{Results}

Under global prediction error (Table~\ref{tab:minimal}), \textsc{Priority} outperforms \textsc{Random} and \textsc{Var-only} but not \textsc{Rotation} or \textsc{Error}. Rotation's guaranteed coverage and error-based allocation's direct feedback both outperform epistemic-gap-guided selection when error is globally observable.

\begin{table}[H]
\centering
\caption{Minimal system ($N\!=\!6$, 2000 runs). Global prediction error, last 50\%. Negative $d$ = \textsc{Priority} is better.}
\label{tab:minimal}
\small
\begin{tabular}{lccccc}
\toprule
Strategy & Error ($\pm$ SD) & vs.\ Pri.\ ($p$) & $d$ & Attn.\ (sw.) \\
\midrule
\textsc{Random} & $0.152 \pm 0.032$ & $< 10^{-6}$ & $-0.15$ & 0.500 \\
\textbf{\textsc{Priority}} & $\mathbf{0.147 \pm 0.030}$ & --- & --- & \textbf{0.534} \\
\textsc{Var-only} & $0.148 \pm 0.031$ & $< 0.001$ & $-0.11$ & 0.509 \\
\textsc{Rotation} & $0.132 \pm 0.024$ & $< 10^{-8}$ & $+0.54$ & 0.500 \\
\textsc{Error} & $0.134 \pm 0.025$ & $< 10^{-8}$ & $+0.48$ & 0.533 \\
\bottomrule
\end{tabular}
\end{table}

\section{Experiment 2: Liminal}

To test whether findings generalize beyond the minimal system, we construct a modular, partially observable environment: 16 variables in 4 modules with heterogeneous dynamics, regime transitions, and inter-variable coupling. Observation budget $b\!=\!2$ per tick. 500 Monte Carlo runs per group.

\subsection{Global Error: Replication of Minimal Results}

Under global prediction error (Table~\ref{tab:liminal_error}), \textsc{Priority} significantly outperforms \textsc{Random} and \textsc{Error}, while matching \textsc{Rotation}. Error-driven allocation collapses ($d\!=\!+0.89$ vs.\ \textsc{Priority}): lacking error signals for unobserved variables, it locks attention onto a small subset, leaving most variables permanently unmonitored. The staleness term in $\pi$ prevents this failure mode.

\begin{table}[H]
\centering
\caption{Liminal ($N\!=\!16$, budget$\!=\!2$, 500 runs). Global prediction error.}
\label{tab:liminal_error}
\small
\begin{tabular}{lcccc}
\toprule
Strategy & Error ($\pm$ SD) & vs.\ Pri.\ ($p$) & $d$ \\
\midrule
\textsc{Random} & $0.267 \pm 0.109$ & $< 10^{-6}$ & $+0.31$ \\
\textsc{Rotation} & $0.236 \pm 0.108$ & $0.961$ & $+0.00$ \\
\textsc{Error} & $0.381 \pm 0.212$ & $< 10^{-8}$ & $+0.89$ \\
\textbf{\textsc{Priority}} & $\mathbf{0.236 \pm 0.093}$ & --- & --- \\
\bottomrule
\end{tabular}
\end{table}

\subsection{The Metric Reversal}

A common assumption in evaluating adaptive systems is that prediction error over all state variables is globally available. Under this assumption, coverage-based strategies are favored, because performance is defined in terms of aggregate error over the entire state space. In partially observable environments, however, error on unobserved variables is not merely unknown to the evaluator---it is unavailable to the agent itself. The central adaptive question is therefore not how accurately the agent predicts everything at once, but how quickly it detects that the environment has changed. By this criterion, our results reverse the standard conclusion: rotation minimizes global error, whereas priority-guided allocation minimizes detection latency, and this advantage grows monotonically with dimensionality.

We define \textit{detection latency} as the number of ticks between a regime switch and the first observation of an affected variable. This metric is more closely aligned with the agent's adaptive problem than global error: it evaluates how quickly environmental change becomes observable through the agent's own sampling process. (Note that computing detection latency still requires an external evaluator to identify regime switch times; however, unlike global error, the underlying quantity---whether a newly observed value deviates from prediction---is available to the agent itself.) We use first observation of an affected variable as the simplest operational proxy; an alternative definition requiring prediction deviation above a threshold would be more conservative but introduces an arbitrary parameter.

\subsection{Detection Speed Scaling}

\begin{figure}[H]
\centering
\includegraphics[width=\columnwidth]{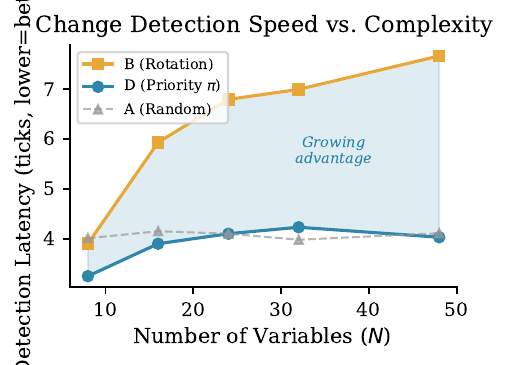}
\caption{Detection latency vs.\ number of variables ($N$), budget$\!=\!1$. Rotation degrades linearly with $N$; priority-guided allocation remains approximately constant. All comparisons $p < 10^{-6}$.}
\label{fig:detection}
\end{figure}

\begin{table}[H]
\centering
\caption{Detection latency (ticks) by strategy and dimensionality. Budget$\!=\!1$, 500 runs. \textsc{Priority} vs.\ \textsc{Rotation}: all $p < 10^{-6}$.}
\label{tab:detection}
\small
\begin{tabular}{rcccccc}
\toprule
$N$ & \textsc{Rand.} & \textsc{Rot.} & \textsc{Pri.} & Gap & $d$ \\
\midrule
8 & 4.01 & 3.90 & \textbf{3.25} & 0.65 & $-0.27$ \\
16 & 4.15 & 5.93 & \textbf{3.90} & 2.02 & $-0.58$ \\
24 & 4.10 & 6.79 & \textbf{4.10} & 2.70 & $-0.72$ \\
32 & 3.98 & 6.99 & \textbf{4.23} & 2.76 & $-0.71$ \\
48 & 4.11 & 7.66 & \textbf{4.03} & 3.63 & $-0.95$ \\
\bottomrule
\end{tabular}
\end{table}

Figure~\ref{fig:detection} and Table~\ref{tab:detection} present the central result. \textsc{Rotation}'s detection latency scales with $N$ (from 3.9 at $N\!=\!8$ to 7.7 at $N\!=\!48$), because it must complete a full cycle before guaranteeing observation of any particular variable. \textsc{Priority} maintains approximately constant detection latency ($\sim$4 ticks) regardless of $N$, because the staleness and surprise terms direct attention toward variables most likely to have changed. The gap widens monotonically: Cohen's $d$ grows from $-0.27$ to $-0.95$.

\subsection{Power Law in Attention Budget}

Fixing $N\!=\!48$ and varying budget $b \in \{1, 2, 4, 8\}$, detection latency follows a power law for both strategies:

\begin{align}
\textsc{Priority:} \quad L &= 4.08 \times b^{-0.55} \quad (R^2 = 0.999) \nonumber \\
\textsc{Rotation:} \quad L &= 8.04 \times b^{-0.40} \quad (R^2 = 0.933) \nonumber
\end{align}

\begin{figure}[H]
\centering
\includegraphics[width=\columnwidth]{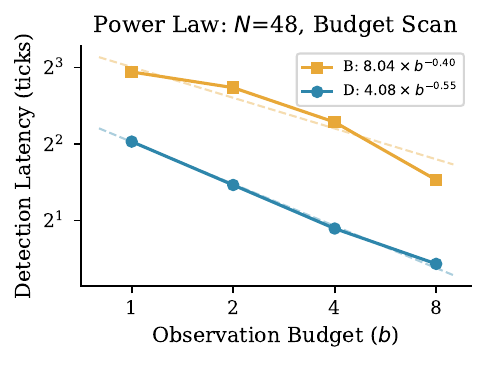}
\caption{Detection latency vs.\ attention budget ($N\!=\!48$, log-log). Priority-guided allocation exhibits a steeper power law exponent (0.55 vs.\ 0.40), extracting greater marginal benefit from additional budget.}
\label{fig:power}
\end{figure}

The steeper exponent for priority-guided allocation (0.55 vs.\ 0.40) means that each additional unit of attention budget yields a proportionally larger improvement in detection speed for the priority function than for rotation. This is because additional observations in a priority-guided system are directed toward high-value targets (via surprise and staleness), whereas additional observations in rotation merely accelerate a fixed cycle.

\section{Experiment 3: Emergent Structure Learning}

\subsection{Motivation}

A limitation of the priority function as specified in Eq.~\ref{eq:priority} is that the staleness term uses a global decay rate $\lambda$, treating all variables identically. In environments with heterogeneous volatility, this is suboptimal: high-volatility variables should be revisited more frequently than low-volatility ones. We ask: if $\lambda$ is made learnable per variable, will the system discover this structure on its own?

\subsection{Design}

We extend the Liminal environment to 16 variables across 4 modules (4 variables each). Modules 0--1 are high-volatility (transition probability $p_{\text{trans}} = 0.15$); Modules 2--3 are low-volatility ($p_{\text{trans}} = 0.02$). The agent is not informed of this structure. Each variable $i$ maintains its own staleness decay rate $\lambda_i$, initialized uniformly at $\lambda_0 = 0.25$, and updated upon each observation of variable $i$ via a surprise-weighted exponential moving average:

\begin{equation}
\lambda_i \leftarrow (1 - \tau)\,\lambda_i + \tau \cdot S_i(t)
\label{eq:lambda_update}
\end{equation}

where $S_i(t) = |x_i - \hat{x}_i| / (\hat{\sigma}_i + \epsilon)$ is the normalized prediction error (surprise) at the time of observation, and $\tau = 0.05$ is the smoothing rate. The intuition is direct: if observations of variable $i$ consistently yield high surprise, $\lambda_i$ increases, causing the staleness term to grow faster when $i$ is unobserved, thereby generating earlier re-observation. If surprise is consistently low, $\lambda_i$ decreases, and the system attends elsewhere. This is not gradient descent through the priority-selection pathway; it is a local heuristic that adjusts temporal urgency based on experienced surprise. Budget $b\!=\!2$, 200 ticks per run, 50 runs. No external labels, no reward signal, no supervision on the volatility structure.

\subsection{Results}

\begin{table}[H]
\centering
\caption{Learned per-variable $\lambda_i$ (50-run mean). Variables 0--7: high volatility ($p_{\text{trans}}\!=\!0.15$); variables 8--15: low volatility ($p_{\text{trans}}\!=\!0.02$). Paired $t$-test on per-run group means: $t(49) = 22.5$, $p < 10^{-6}$.}
\label{tab:lambda}
\small
\begin{tabular}{cccc|cccc}
\toprule
Var & Mod & Vol.\ & $\lambda_i$ & Var & Mod & Vol.\ & $\lambda_i$ \\
\midrule
0 & 0 & Hi & 0.296 & 8 & 2 & Lo & 0.198 \\
1 & 0 & Hi & 0.292 & 9 & 2 & Lo & 0.200 \\
2 & 0 & Hi & 0.291 & 10 & 2 & Lo & 0.186 \\
3 & 0 & Hi & 0.300 & 11 & 2 & Lo & 0.199 \\
4 & 1 & Hi & 0.284 & 12 & 3 & Lo & 0.213 \\
5 & 1 & Hi & 0.283 & 13 & 3 & Lo & 0.206 \\
6 & 1 & Hi & 0.280 & 14 & 3 & Lo & 0.209 \\
7 & 1 & Hi & 0.284 & 15 & 3 & Lo & 0.202 \\
\midrule
\multicolumn{3}{c}{High mean} & \textbf{0.289} & \multicolumn{3}{c}{Low mean} & \textbf{0.202} \\
\bottomrule
\end{tabular}
\end{table}

\begin{figure}[H]
\centering
\includegraphics[width=\columnwidth]{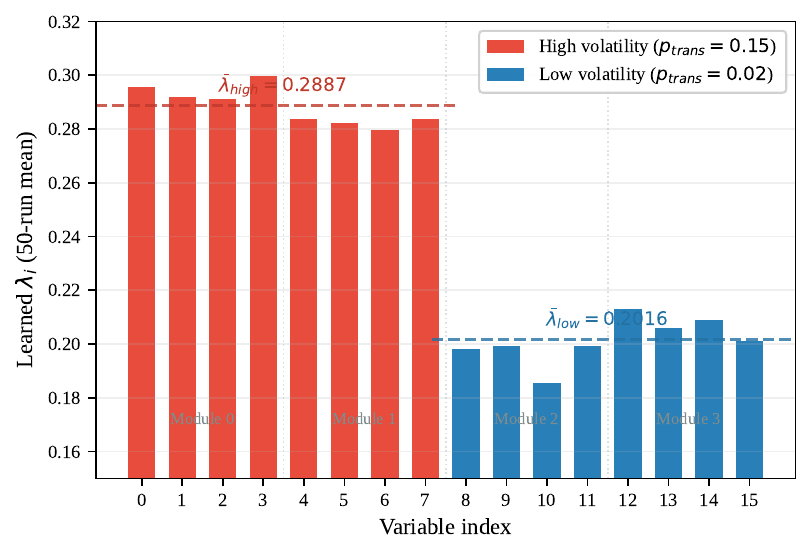}
\caption{Learned $\lambda_i$ per variable (50-run mean). Red: high-volatility modules ($p_{\text{trans}}\!=\!0.15$); blue: low-volatility modules ($p_{\text{trans}}\!=\!0.02$). The system spontaneously differentiates volatility structure without supervision ($t = 22.5$, $p < 10^{-6}$).}
\label{fig:lambda}
\end{figure}

Table~\ref{tab:lambda} and Figure~\ref{fig:lambda} present the central result. The learned $\lambda_i$ values cleanly separate into two clusters corresponding to the true volatility structure: high-volatility variables converge to $\bar{\lambda}_{\text{high}} = 0.2886$ ($\pm 0.007$), while low-volatility variables converge to $\bar{\lambda}_{\text{low}} = 0.2016$ ($\pm 0.009$). To avoid pseudo-replication (variables within a run are not independent), we compute the mean $\lambda$ for high-volatility and low-volatility groups within each run, yielding 50 paired observations. A paired $t$-test on these per-run means confirms the separation: $t(49) = 22.5$, $p < 10^{-6}$.

This result was not programmed. No label told the system which variables are volatile. No reward signal reinforced correct $\lambda$ values. The closed loop---priority $\to$ observation $\to$ prediction error $\to$ $\lambda$ update $\to$ changed priority---self-organized into a representation of the environment's volatility structure. The system learned \textit{where change happens} by attending to its own epistemic gaps. We note that the current experiment demonstrates recovery of a coarse binary partition (high vs.\ low volatility with a 7.5$\times$ gap in transition probability); whether the mechanism recovers finer-grained structure under smaller volatility differences remains an open question.

\section{Discussion}

\subsection{The Evaluation Assumption}

Our results expose a methodological assumption that pervades evaluation of adaptive systems: that prediction error is globally observable. This assumption is inherited from supervised learning, where loss is computed over labeled data available in full. In partially observable environments, this assumption breaks: the agent cannot know how wrong it is about variables it has not observed. Defining performance as global prediction error implicitly rewards a capacity the agent does not have---omniscient self-evaluation---and thereby favors strategies optimized for a world in which loss is always visible.

When we replace this metric with one available to the agent---how quickly it detects environmental change through its own observations---the ranking reverses. This is not a matter of choosing a metric favorable to our method; it is a matter of choosing a metric consistent with the problem setting. In partially observable worlds, the first-order adaptive challenge is not ``minimize error everywhere'' but ``discover where error has appeared.''

\subsection{Why Staleness Matters}

Error-driven allocation collapses in the Liminal environment ($d\!=\!+0.89$ vs.\ \textsc{Priority}) because it has no signal for unobserved variables. It can only respond to errors it has already seen, creating a feedback loop that concentrates attention on a shrinking subset. The staleness term $(1 - e^{-\lambda \Delta t_i})$ breaks this loop by generating priority from temporal reasoning alone: ``I have not observed this variable in a while; it may have changed.'' This requires no external feedback, no error signal, no reward---only an internal clock and an assumption of environmental non-stationarity.

\subsection{Implications for Cognitive Architecture}

If attentional priorities can be generated endogenously from epistemic gaps, then priority generation constitutes a distinct computational layer between world model and policy. This suggests a route toward endogenous goal formation in more general architectures, where the priority function could drive not just observation but action selection, exploration planning, and resource allocation.

The power law result suggests a deeper principle: in attention-limited systems, the \textit{structure} of attention allocation matters more than its \textit{quantity}. Doubling the observation budget improves priority-guided detection by a factor of $2^{0.55} = 1.46$, but rotation by only $2^{0.40} = 1.32$. Structured allocation amplifies the value of scarce resources.

\subsection{Spontaneous Structure Recovery}

The learned-$\lambda$ result (Experiment 3) goes beyond parameter tuning. The system was given no information about which variables are volatile, yet it recovered the ground-truth volatility partition through closed-loop interaction alone. This is an instance of \textit{unsupervised environment structure learning} driven entirely by epistemic gaps: the priority function creates an observation pattern; observations generate prediction errors; errors update $\lambda_i$; updated $\lambda_i$ reshape the priority function. No component in this loop has access to ground truth. The structure emerges from the loop itself.

This suggests a broader hypothesis: systems equipped with a world model, an uncertainty function, and an endogenous goal-generation mechanism may be sufficient to produce adaptive behavior organization without external reward. The present work provides empirical evidence supporting this hypothesis in the restricted setting of attention allocation under partial observability. Whether the same principle extends to action selection, planning, and resource allocation in richer environments is a question we leave to future work.

\subsection{Limitations}

Experiment 3 addresses the global-$\lambda$ limitation identified in earlier versions, but introduces new questions. The surprise-weighted update rule is a heuristic; more principled alternatives (e.g., Bayesian model selection over discrete volatility categories, or online changepoint detection) remain unexplored. Detection latency, while more agent-centric than global error, still requires post-hoc identification of regime switches by an external evaluator; a fully intrinsic metric remains future work. All experiments use Bayesian agents with known model class; extension to agents with learned world models (e.g., neural networks) is a natural but non-trivial next step. The current work provides empirical evidence for the sufficiency of epistemic gaps in attention-allocation tasks; whether this mechanism extends to action selection and planning in richer environments is an open question.

\section{Conclusion}

We have presented Telogenesis, a framework in which attentional priorities emerge from epistemic gaps. The priority function $\pi_i(t)$ unifies ignorance, surprise, and staleness into a mechanism for observation allocation. The central finding is a metric-dependent reversal: under global error, coverage wins; under detection latency, priority-guided allocation wins, with advantage growing monotonically with complexity and following a power law in attention budget (exponent 0.55 vs.\ 0.40).

When the staleness decay rate is made learnable, the system spontaneously recovers the volatility structure of the environment without any external supervision ($t(49) = 22.5$, $p < 10^{-6}$). This provides empirical evidence that epistemic gaps are sufficient not only for adaptive priority generation, but for unsupervised discovery of latent environmental structure.

The first-order advantage of endogenous priority is not uniform coverage, but accelerated change detection under attentional scarcity. Goal is all u need.

\bigskip
\textit{Written before the proof, kept for after.}

\small
\bibliographystyle{apalike}

\end{document}